\documentclass{SCIS2023}

\usepackage[ruled, vlined, linesnumbered]{algorithm2e}

\usepackage{colortbl}
\usepackage{xcolor}
\usepackage{tabularx, makecell, multirow} 
\definecolor{mygray}{gray}{.9}
\begin{document}
\ArticleType{RESEARCH PAPER}
\Year{X}
\Month{X}
\Vol{X}
\No{X}
\DOI{10.1007/s11432-XXX-XXXX-X}
\ArtNo{}
\ReceiveDate{16 July 2023}
\ReviseDate{}
\AcceptDate{11 June 2024}
\OnlineDate{}

\title{AsyCo: An Asymmetric Dual-task Co-training Model for Partial-label Learning}{AsyCo: An Asymmetric Dual-task Co-training Model for Partial-label Learning}

\author[1]{Beibei Li }{}
\author[2,3]{Yiyuan Zheng }{}
\author[2,3]{Beihong Jin }{}
\author[1]{Tao Xiang }{{txiang@cqu.edu.cn}}
\author[4]{Haobo Wang }{{wanghaobo@zju.edu.cn}}
\author[5]{Lei Feng }{}

\AuthorMark{Beibei Li}

\AuthorCitation{Beibei Li, Yiyuan Zheng, Beihong Jin, et al.}


\address[1]{College of Computer Science, Chongqing University, Chongqing {\rm 400044}, China}
\address[2]{State Key Laboratory of Computer Science, Institute of Software, Chinese Academy of Sciences, Beijing {\rm 100044}, China}
\address[3]{University of Chinese Academy of Sciences, Beijing {\rm 100044}, China}
\address[4]{School of Software Technology, Zhejiang University, Hangzhou {\rm 310027}, China}
\address[5]{Information Systems Technology and Design Pillar, Singapore University of Technology and Design, Singapore {\rm 487372}}

\abstract{Partial-Label Learning (PLL) is a typical problem of weakly supervised learning, where each training instance is annotated with a set of candidate labels. Self-training PLL models achieve state-of-the-art performance but suffer from error accumulation problem caused by mistakenly disambiguated instances. Although co-training can alleviate this issue by training two networks simultaneously and allowing them to interact with each other, most existing co-training methods train two structurally identical networks with the same task, i.e., are symmetric, rendering it insufficient for them to correct each other due to their similar limitations. Therefore, in this paper, we propose an asymmetric dual-task co-training PLL model called AsyCo, which forces its two networks, i.e., a disambiguation network and an auxiliary network, to learn from different views explicitly by optimizing distinct tasks. Specifically, the disambiguation network is trained with self-training PLL task to learn label confidence, while the auxiliary network is trained in a supervised learning paradigm to learn from the noisy pairwise similarity labels that are constructed according to the learned label confidence. Finally, the error accumulation problem is mitigated via information distillation and confidence refinement. Extensive experiments on both uniform and instance-dependent partially labeled datasets demonstrate the effectiveness of AsyCo. The code is available at https://github.com/libeibeics/AsyCo.}

\keywords{machine learning, weakly supervised learning, partial-label learning, co-training models, candidate label sets}

\maketitle

\section{Introduction}
\begin{figure}[!t] 
    \centering 
    \includegraphics[width=0.5\textwidth]{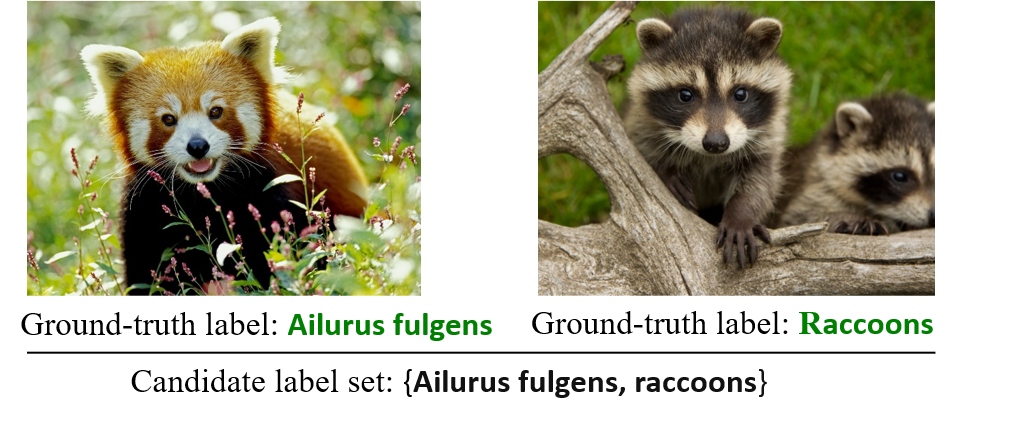}
    \caption{Two examples of partially labeled instances. Due to the visual similarity between Ailurus fulgens and raccoons, the two images are both annotated with 'Ailurus fulgens' and 'raccoons'.}
    \label{fig:example} 
\end{figure}

Training deep neural networks via supervised learning requires massive accurately-annotated data, which are, however, expensive to be collected. To overcome this problem, weakly supervised learning~\cite{1,point_annotation,semantic_seg,multilabel_learning} has been widely studied in recent years. Partial-Label Learning (PLL)~\cite{2,robust_svm} is a typical type of weakly supervised learning with inaccurate supervision, which assumes that each training instance is annotated with a candidate label set that contains the ground-truth label. As shown in Fig. \ref{fig:example}, the visual resemblance between raccoons and Ailurus fulgens makes it challenging for annotators to confidently pinpoint the exact animal depicted in the images. As a result, they assign multiple candidate labels to each image, leading to  partially labeled instances. Since label ambiguity is pervasive in data annotations, PLL has been widely applied in various real-world applications, such as automatic image annotation~\cite{3} and multimedia content analysis~\cite{4}.

Recent research on PLL has primarily  concentrated on identification-based methods, which regard the ground-truth label as a latent variable and try to recognize the ground-truth label by conducting label disambiguation. To this end, various techniques have been employed, such as maximum margin~\cite{5}, graph models~\cite{6,7,8,9}, expectation-maximum algorithm~\cite{10}, contrastive learning~\cite{11}, and consistency regularization~\cite{12}.  Among these, self-training deep models~\cite{13,14,15} have emerged as a promising approach, {  which learn label confidence vectors and train the models with them iteratively, } and achieved state-of-the-art performance~\cite{12}.

However, self-training PLL models suffer from a problem of error accumulation, because complicated instances are difficult to classify and easy to be mistakenly disambiguated, which could further mislead the model with false positive labels and causes performance degradation. The co-training strategy~\cite{16,17}, which trains two networks simultaneously and makes them interact with each other, is a feasible solution to mitigate error accumulation.  While the co-training strategy has been extensively explored in Noisy Label Learning (NLL) ~\cite{18,19,20}, its usage in PLL remains understudied. Recently, Yao et al.~\cite{21} proposed a novel approach called NCPD for PLL based on co-training. NCPD transforms partially labeled datasets into highly noisy datasets via data duplication and adopts a typical NLL method called co-teaching~\cite{20}. However, NCPD not only causes extremely high time and space complexity but also obtains limited performance.

Moreover, the majority of existing co-training models, including NCPD for PLL, {  are symmetric, i.e., their two branches of networks have the same structure and are trained with the same input data and loss functions. They} assume that different parameter initializations enable the two structurally identical networks trained with the same task to obtain distinct capabilities so that they are able to provide mutual corrections for each other. Nevertheless, being trained {   in a symmetric paradigm} makes the two networks fall into the same limitations easily, e.g., both of them are hard to recognize complicated instances correctly. Consequently, they are easier to reach a consensus and cannot correct errors for each other effectively. 

Therefore, we argue that training the two networks {  in asymmetric paradigm by constructing different structure for each branch of network carefully or training them with distinct input data or loss functions from different views,} can explicitly enable them to {  capture disparate information} and enhance the possibility to get complementary capabilities, which benefits error correction. Intuitively, in PLL, the partially labeled dataset can be transformed into an exactly labeled dataset by annotating each instance with a pseudo label corresponding to the maximum confidence. Then, the two networks can be trained via a partial-label learning task and a supervised learning task, respectively. However, training with the generated exactly labeled dataset is challenging since mistakenly disambiguated instances could be annotated with noisy class labels, which could be harmful to model learning. Fortunately, according to Wu et al.~\cite{22}, under mild conditions, when the number of classes $c\leq 8$, the noise rate of noisy pairwise similarity labels, i.e., labels indicating whether or not two instances belong to the same class, is lower than that of the noisy class labels. Thus, converting the noisy class labels into noisy similarity labels can reduce the influence of noisy class labels.

{  In the light of the above motivations,} we propose an asymmetric dual-task co-training PLL model AsyCo. AsyCo comprises two networks that share identical structures but are trained by distinct tasks. The first network is designed as a disambiguation network that focuses on resolving label ambiguity and is trained using a self-training method, for the PLL task. According to its learned confidence, we generate pseudo class labels for instances by annotating each training instance with the most confident label, and further transform the noisy pseudo class labels into noisy pairwise similarity labels, of which the noise rates are much lower. Then, the second network, referred to as an auxiliary network, is then trained using the generated noisy similarity labels in a supervised learning paradigm. { With the clarified information provided by the disambiguation network, the auxiliary network is trained with higher-quality data, improving the probability of classifying complicated instances correctly. Besides, the auxiliary network utilizes different data and loss function from the disambiguation network for training and makes it easier to obtain the complementary classification capabilities of the disambiguation network. Therefore, } we leverage the prediction of the auxiliary network to conduct error correction for the disambiguation network and boost classification accuracy finally.

Overall, our contributions can be summarized as follows:

1). We explore asymmetric co-training for PLL and propose a novel deep dual-task PLL model AsyCo that trains two structurally identical networks with distinct tasks collaboratively.

2). As an integral part of AsyCo, an effective supervised learning auxiliary network is proposed, which utilizes the pseudo labels identified by the disambiguation network for training and mitigates the error accumulation problem via distillation and refinement in turn.

3). Extensive experimental results on benchmark datasets demonstrate the superior performance of AsyCo on both uniform and instance-dependent partially labeled data.


The rest of the paper is organized as follows. First, some necessary preliminary knowledge of PLL is  illustrated. Next, we present the proposed AsyCo model in Section \ref{sec:ourmodel} and  report  experiments in Section \ref{sec:exp}, respectively. Then, related work is briefly reviewed. Lastly, we conclude this paper. 

\section{Preliminaries}\label{sec:prelimi}
\subsection{Problem Settings}

\noindent We denote $\mathcal{X}\subset\mathbb{R}^d$ as the $d$-dimensional feature space and $\mathcal{Y}\subset\{1,2,\ldots,m\}$ as the label space. In partially labeled datasets, each training instance $\bm{x}_i\in\mathcal{X}$ is labeled with a candidate label set ${Y}_i\subset\mathcal{Y}$ that contains the ground-truth label $y_i$. Our goal is to learn a multi-class classifier $f\left(\cdot\right)$ on partially label dataset $\mathcal{D}=\{(\bm{x}_i, Y_i)|1\leq i \leq n\}$. We use $p_{ik} = f_k(\bm{x}_i)$ to denote the predicted probability of classifier $f\left(\cdot\right)$ on label $k$ given instance $\bm{x}_i$. 

Note that for non-structural instances, such as images and text,  their $d$-dimensional features are typically extracted using Deep Neural Network(DNN)-based encoders from their raw feature. For instance, the feature encoder for images is commonly constructed based on convolutional neural networks like LeNet, ResNet, or WideResNet, etc.

\subsection{Classifier-consistent PLL Loss}

\noindent The Classifier-Consistent (CC) PLL loss~\cite{14} assumes each candidate label set is uniformly sampled and is presented as follows:
\begin{gather}
L_{\mathrm{cc}}\left(\bm{x}_i\right)=-\log \Big(\sum\nolimits_{k \in Y_i} p_{i k}\Big).
\end{gather}
Minimizing the CC loss is equivalent to maximizing the sum of the classification probabilities of all the candidate labels while minimizing the sum of the classification probabilities of non-candidate labels. 

\subsection{Risk-consistent PLL Loss}

\noindent The above classifier-consistent PLL loss~\cite{14} assigns the same weight to each candidate label and makes the ground-truth label easily overwhelmed by other false positive labels. Thus, a Risk-Consistent (RC) loss based on the importance-weighting strategy was proposed ~\cite{14}. By leveraging the widely-used categorical cross entropy loss as the basic classification loss, the risk-consistent PLL loss is formulated as follows:
\begin{gather}
L_{\mathrm{r c}}\left(\bm{x}_i\right)=-\sum\nolimits_{k=1}^m c_{ik}\log p_{i k},   \quad c_{ik} = \frac{p\left(y_i=k \mid \bm{x}_i\right)}{\sum\nolimits_{j \in Y_i} p\left(y_i=j \mid \bm{x}_i\right)},    
\end{gather}
where $p\left(y_i=k \mid \bm{x}_i\right)$ represents the probability that instance $\bm{x}_i$ belongs to category $k$. Actually, $c_{i k}$ implies how confident the probability of falling into category $k$ is, thus, a confidence vector can be formed as $\left[c_{i 0}, c_{i 1}, \ldots, c_{i m}\right]$. Since $p\left(y_i=k \mid \bm{x}_i\right)$ is not accessible from the given data, it is approximated by the classification probability, shown as,
\begin{align}\label{eq:rc_probability_k}
p\left(y_i=k \mid \bm{x}_i\right) &= \left\{\begin{array}{cl}
f_k\left(\bm{x}_i\right) & \text { if } k \in Y_i, \\
0 & \text { otherwise }.
\end{array}\right.
\end{align}

By calculating $p\left(y_i=k \mid \bm{x}_i\right)$ as above, the RC PLL loss trains models in a self-training manner.

\section{The Proposed Model} \label{sec:ourmodel}

\begin{figure*} [!t]
    \centering 
    \includegraphics[width=0.9\textwidth]{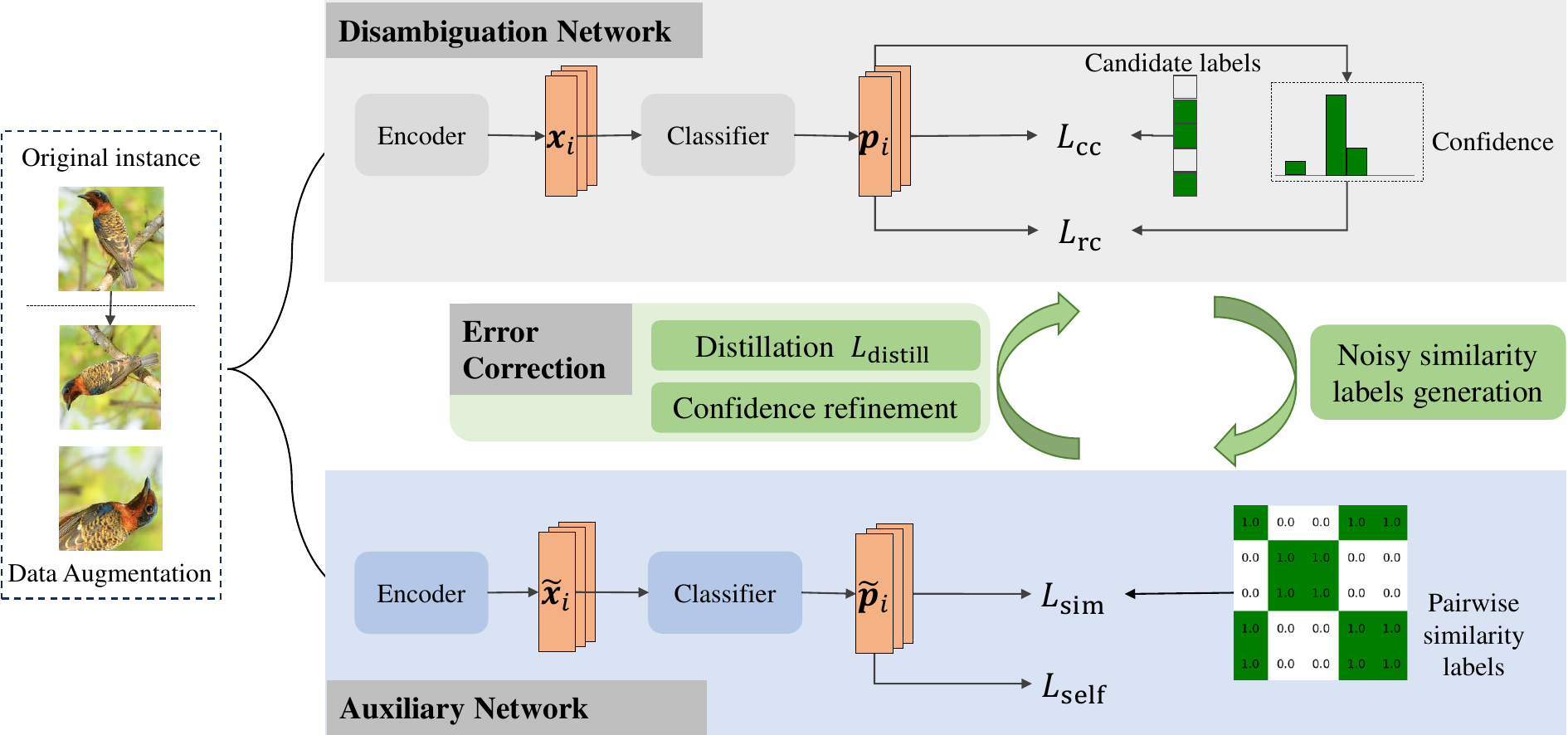}
    \caption{Architecture of AsyCo (in the training phase). AsyCo comprises two networks with identical structures, namely the disambiguation network and the auxiliary network. The former is responsible for resolving label ambiguities and learning label confidence, while the latter is trained by pairwise similarity labels constructed according to the learned confidence. Besides, the auxiliary network facilitates error correction for the disambiguation network through information distillation and confidence refinement, thereby mitigating the error accumulation problem.}
    \label{fig:model} 
\end{figure*}

\noindent As shown in Fig. \ref{fig:model}, our model is composed of a disambiguation network, an auxiliary network and an error correction module. The two networks have identical structures but are trained with different tasks and loss functions, leading to discrepancies in their parameters and capabilities. Specifically, the disambiguation network undergoes training using PLL losses and obtains the confidence of each label in the candidate label set, while the auxiliary network leverages low-noise pairwise similarity labels generated according to the learned label confidence and is trained via supervised learning losses. Finally, the auxiliary network addresses the issue of error accumulation in the disambiguation network through information distillation and confidence refinement. To facilitate understanding, the notations set in AsyCo are summarized in Table \ref{tab:notation}.

\begin{table}[t]
\footnotesize
\caption{Notations for AsyCo}
\label{tab:notation}
\begin{center}
\begin{tabular}{l|l}
\toprule
Notation    & Description\\ 
\midrule
$\mathcal{X}\subset\mathbb{R}^d$ & $d$-dimensional feature space\\
$\mathcal{Y}\subset\{1,2,\ldots,m\}$ & Label space\\
$\bm{x}_i\in\mathcal{X}, \tilde{\bm{x}}_i\in\mathcal{X}$ & Features of the $i$th instance in the disambiguation network and the auxiliary network, respectively\\
$Y_i\subset\mathcal{Y}$ & Candidate label set of the instance $\bm{x}_i$\\
$\mathcal{D}=\{(\bm{x}_i, Y_i)|1\leq i \leq n\}$ & Partially labeled dataset\\
$\mathcal{A}(\bm{x}_i)=\{\bm{x}_i, \bm{x}_i^\prime, \bm{x}_i^{\prime\prime}\}$ & Instance set containing the original instance $\bm{x}_i$ and its augmentations\\
$f\left(\cdot\right)$ & Classifier that takes instance feature as input\\
$\bm{p}_i, \bm{p}_i^\prime, \bm{p}_i^{\prime\prime}, \bm{c}_i, \bm{c}_i^\prime, \bm{c}_i^{\prime\prime}$ & Predicted probabilities and label confidence vetors of the instances in $\mathcal{A}(\bm{x}_i)=\{\bm{x}_i, \bm{x}_i^\prime, \bm{x}_i^{\prime\prime}\}$ \\
$\bm{w}_{i}=[w_{i0}, w_{i1}, \ldots, w_{im}]$ & The comprehensive label confidence vector of the instances in  $\bm{x}_i$ that integrates $\bm{c}_i, \bm{c}_i^\prime$ and $\bm{c}_i^{\prime\prime}$ \\
$\gamma(t)$ & Non-decreasing factor balancing CC loss and RC loss\\
$k^{\prime}$ &  Pseudo label of instance $\bm{x}_i$ generated according to label confidence vector $\bm{w}_i$\\
\midrule
$\tilde{f}\left(\cdot\right)$ & Classifier in the auxiliary network\\
$\mathcal{A}(\tilde{\bm{x}}_i)=\{\tilde{\bm{x}}_i, \tilde{\bm{x}}_i^\prime, \tilde{\bm{x}}_i^{\prime\prime}\}$ & Instance set containing the original instance $\tilde{\bm{x}}_i$ and its augmentations in the auxiliary network\\
$\bm{\tilde{p}}_i,\bm{\tilde{p}}_i^\prime,\bm{\tilde{p}}_i^{\prime\prime}$ & Predicted probability of instances in $\mathcal{A}(\tilde{\bm{x}}_i)$ calculated by the auxiliary network \\
$\tilde{\bm{w}}_{i}$ & The comprehensive label confidence vector of the instance $\tilde{\bm{x}}_i$ that integrates $\tilde{\bm{c}}_i, \tilde{\bm{c}}_i^\prime$ and $\tilde{\bm{c}}_i^{\prime\prime}$ \\
$s_{ij} \in\{0,1\}$ & Generated similarity label between the  instance $\tilde{\bm{x}}_i$ and the instance $\tilde{\bm{x}}_j$\\
$\mathcal{D}_{s i m}=\left\{\langle \tilde{\bm{x}}_i, \tilde{\bm{x}}_j\rangle, s_{i j}\right\}$ & Generated similarity dataset, where $\langle \tilde{\bm{x}}_i, \tilde{\bm{x}}_j\rangle \in \mathcal{X} \times \mathcal{X}$\\
$\mu(t)$ & Non-decreasing refinement factor of the label confidence calculated by the auxiliary network\\
$\hat{\bm{w}}_{i}$ & The refined label confidence vector of instance $\bm{x}_i$ \\
$\tau, T, \lambda$ & Hyper-parameters\\
\bottomrule
\end{tabular}
\end{center}
\end{table}

\subsection{Disambiguation Network}

\noindent Given an instance $\bm{x}_i$, a classifier is proposed to compute classification logits using a Multi-Layer Perceptron (MLP). The classifier further calculates the classification probability $\bm{p}_i \in \mathbb{R}^m$ by applying the softmax function. Specifically, the $k$th element of $\bm{p}_i$, which represents the probability that the instance $\bm{x}_i$ is classified into the $k$th category, is determined as follows:
\begin{gather}\label{eq:prob_ik}
p_{i k}=p\left(y_i=k \mid \bm{x}_i\right) =f\left(\bm{x}_i\right)=\frac{\exp \left(\operatorname{MLP}_k\left(\bm{x}_i\right) / \tau\right)}{\sum_j \exp \left(\operatorname{MLP}_j\left(\bm{x}_i\right) / \tau\right)},
\end{gather}
where $\operatorname{MLP}_k (\bm{x}_i)$ denotes the classification logit for classifying instance $\bm{x}_i$ into label $k$. Additionally, $\tau$ serves as  a temperature parameter, wherein a higher value of $\tau$ results in smoother classification probabilities. 

Data augmentation, which generates richer and harder instances by making slight modifications to the original instances, benefits classification performance. Following DPLL~\cite{12}, we apply two augmentation methods, i.e., Autoaugment~\cite{23} and Cutout~\cite{24}, to generate two augmentated views for each instance, which are denoted as $\bm{x}_i^{\prime}=\operatorname{Aug} _1\left(\bm{x}_i\right)$ and $\bm{x}_i^{\prime \prime}=\operatorname{Aug}_2\left(\bm{x}_i\right)$. The original instance and the augmented instances of $\bm{x}_i$ form a instance set $\mathcal{A}\left(\bm{x}_i\right)$, i.e., $\mathcal{A}\left(\bm{x}_i\right)=\left\{\bm{x}_i, \bm{x}_i^{\prime}, \bm{x}_i^{\prime \prime}\right\}$. Besides, $\bm{x}_i^{\prime}, \bm{x}_i^{\prime \prime}$ and $\bm{p}_i^{\prime}, \bm{p}_i^{\prime \prime}$ denote the features and classification probabilities of the augmented instances, respectively. According to Equation \ref{eq:cik}, the label confidence vectors of the origin instance $\bm{x}_i$ and its augmentations $\bm{c}_i^{\prime}, \bm{c}_i^{\prime \prime}$ can be calculated successively, which are denoted as $\bm{c}_i, \bm{c}_i^\prime, \bm{c}_i^{\prime\prime}$, respectively. 

We extend the original CC and RC losses to accommodate the augmented instances, and train the disambiguation network with the enhanced losses. Specifically, with data augmentation, the CC loss of instance $\bm{x}_i$ can be computed as follows:
\begin{gather}
L_{\mathrm{cc}}\left(\bm{x}_i\right)=-\frac{1}{\left|\mathcal{A}\left(\bm{x}_i\right)\right|} \sum\nolimits_{\bm{x}_i^* \in\mathcal{A}\left(\bm{x}_i\right)} \log \left(\sum\nolimits_{k \in Y_i} p_{i k}^*\right),
\end{gather}
where $|\mathcal{A}(\bm{x}_i)|$ represents the cardinality of $\mathcal{A}(\bm{x}_i)$.

As for $\mathrm{RC}$ loss, we denote the confidence vector of instance $\bm{x}^*_i \in \mathcal{A}(\bm{x}_i)$ as $\bm{c}_i^*$ and integrate the confidence vectors of all the instances in $\mathcal{A}\left(\bm{x}_i\right)$ to calculate a comprehensive label confidence vector $\bm{w}_i$, which is more accurate and robust. In detail, the confidence corresponding to class $k$ for instance $\bm{x}_i$ can be calculated as follows:
\begin{gather}\label{eq:cik}
w_{i k}=\frac{\left(\prod_{\bm{x}^*_i \in \mathcal{A}(\bm{x}_i)} c_{i k}^*\right)^{\frac{1}{|\mathcal{A}(\bm{x}_i)|}}}{\sum_{j \in Y_i}\left(\prod_{\bm{x}^*_i \in \mathcal{A}(\bm{x}_i)} c_{i j}^*\right)^{\frac{1}{|\mathcal{A}(\bm{x}_i)|}}}. 
\end{gather}

Then, the data augmentation enhanced RC loss of instance $\bm{x}_i$, which is a self-training loss, is as follows:
\begin{gather}\label{eq:lrc}
L_{\mathrm{rc}}\left(\bm{x}_i\right)=-\frac{1}{\left|\mathcal{A}\left(\bm{x}_i\right)\right|} \sum\nolimits_{\bm{x}_i^* \in\mathcal{A}\left(\bm{x}_i\right)} \sum\nolimits_{k \in Y_i}  w_{i k} \log p_{i k}^*.    
\end{gather}
Minimizing the data augmentation enhanced RC loss simultaneously drives the classification probabilities of both the original and augmented instances closer to the confidence vector $ {[}\bm{w}_{i} = {w_{i1},w_{i2},\ldots, w_{im}}{]}$. This process encourages the network's output to be invariant to minor changes made in the feature space, thereby implicitly extracting self-supervised information.

Finally, the disambiguation network is trained by the following constructed loss:
\begin{gather}\label{eq:ldisam}
    L_{\mathrm{disam}}\left(\bm{x}_i\right)= L_{\mathrm{cc}}\left(\bm{x}_i\right) + \gamma(t)L_{\mathrm{rc}}\left(\bm{x}_i\right).
\end{gather}

We introduce a coefficient $\lambda$ for the RC loss to address the potential instability of learned confidence vectors in the early stages of training. This coefficient serves as a non-decreasing balancing function that gradually increases over the training epoch number $t$., i.e.,
\begin{gather}\label{eq:gamma_t}
\gamma(t)=\min \left\{\frac{t}{T} \lambda, \lambda\right\},
\end{gather}
where $\lambda$ and $T$ are hyper-parameters.

\subsection{Auxiliary Network}

\begin{figure}[!t] 
    \centering 
    \includegraphics[width=0.6\textwidth]{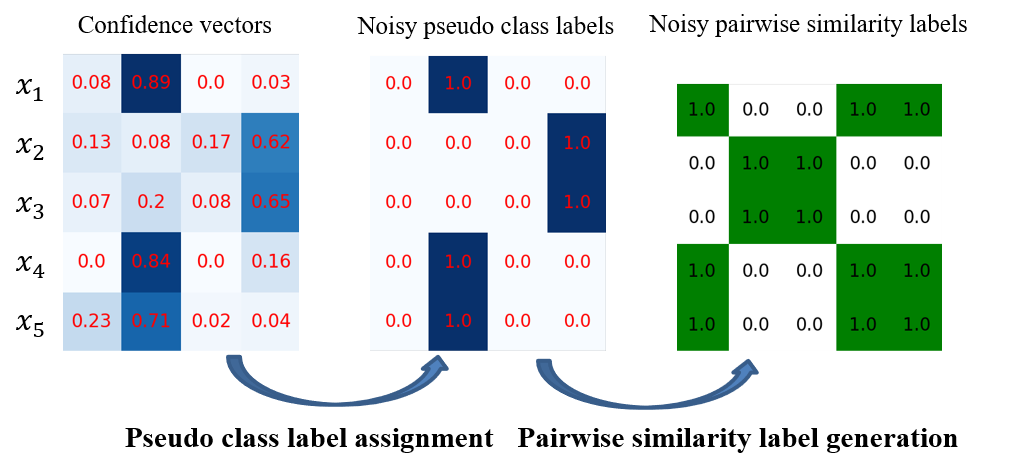}
    \caption{An example of generating noisy pairwise similarity labels according to confidence vectors.}
    \label{fig:pic_similarity} 
\end{figure}

\noindent The auxiliary network has the same structure as the disambiguation network, but their parameters are trained differently. To symbolically differentiate the components and variables of these two networks, the symbols corresponding to the auxiliary network are augmented with tilde symbols. For example, in the auxiliary network, the classifier, the feature, the predicted classification probabilities and the final label confidence are denoted as $\tilde{f}(\cdot)$, $\tilde{\bm{x}}_i$, $\tilde{\bm{p}}_i$ and $\tilde{\bm{w}}_i$ respectively.

As the disambiguation network undergoes training, the precision of confidence vector continually improves. For a certain portion of instances, their ground-truth labels are identified by the maximum confidence value. This motivates us to assign an exact pseudo class label to each instance according to its label confidence, thereby sufficiently utilizing this clarified information to enhance model training. As shown in Fig. \ref{fig:pic_similarity}, assuming instance $\bm{x}_i$ gets the largest confidence on the $k^{\prime}$th label, i.e., $k^{\prime}=\operatorname{argmax}_k\left\{w_{ik} \mid y_{ik}\in Y_i\right\}$, we generate a one-hot label vector for it, of which the $k^{\prime}$th element is 1, other elements are 0. Consequently, we obtain a new dataset where each instance is annotated with an exact class label. Due to that some instances may be mistakenly disambiguated and annotated, there are some noisy data in the generated dataset.

Refer to the work of Wu et al.~\cite{22}, the noise rates of the pairwise similarity labels are lower than that of the intermediate noisy class labels in most practical scenarios. Therefore, we transform the pseudo class labels into pairwise similarity labels, and utilize the resulting similarity dataset to train the auxiliary network. Specifically, for each pair of instances, if they share the same pseudo class label, we assign a similarity label of 1 to them; otherwise, we assign a similarity label of 0. The generated similarity dataset is denoted as $\mathcal{D}_{s i m}=\left\{\langle \tilde{\bm{x}}_i, \tilde{\bm{x}}_j\rangle, s_{i j}\right\}$, where $\langle \tilde{\bm{x}}_i, \tilde{\bm{x}}_j\rangle \in \mathcal{X} \times \mathcal{X}, s_{i j} \in\{0,1\}$. 

For pairs of instances with a similarity label of 1, it is expected that their predicted classification probabilities exhibit high similarity. To capture this inter-instance relationship, we propose a binary cross-entropy loss function as Equation \ref{eq:lsim}. It should be noted that both $\tilde{\bm{p}}_{i}$ and $\tilde{\bm{p}}_{j}$ are classification probabilities normalized by the softmax function, thus, $0\leq\tilde{\bm{p}}_i^T \tilde{\bm{p}}_j\leq 1$.
\begin{gather}\label{eq:lsim}
L_{\mathrm{sim}}\left(\tilde{\bm{x}}_i, \tilde{\bm{x}}_j, s_{i j}\right)=\frac{1}{\left|\mathcal{A}(\tilde{\bm{x}}_i)\right|}\sum\nolimits_{\tilde{\bm{x}}^*_i \in \mathcal{A}(\tilde{\bm{x}}_i)}-s_{i j} \log \left(\tilde{\bm{p}}_i^T \tilde{\bm{p}}_j\right)-\left(1-s_{i j}\right) \log \left(1-\tilde{\bm{p}}_i^T \tilde{\bm{p}}_j\right).   
\end{gather}

Furthermore, in order to keep the consistency between the original instances and their augmentations, we construct the following cross entropy-based loss to learn self-supervised information:
\begin{gather}
    L_{\mathrm{ssl}}\left(\tilde{\bm{x}}_i\right) = -\frac{1}{2}\sum\nolimits_{\tilde{\bm{x}}_i^* \in\left\{\tilde{\bm{x}}_i^{\prime}, \tilde{\bm{x}}_i^{\prime \prime}\right\}} \sum\nolimits_{y_{ik} \in Y_i}\left(\operatorname{stop-grad}(\tilde{p}_{ik}) \log\tilde{p}_{ik}^*\right).
\end{gather}
where $\operatorname{stop-grad}(\tilde{p}_{ik})$ denotes that we stop the gradients of  $\tilde{p}_{ik}$ in $L_{\mathrm{ssl}}$ during back-propagation. 

Finally, the overall loss to train the auxiliary network is as follows:
\begin{gather}
    L_{\mathrm{aux}}(\tilde{\bm{x}}_i) = L_{\mathrm{ssl}}(\tilde{\bm{x}}_i) + \gamma(t)\Big(\frac{1}{n}\sum\nolimits_{j=0}^nL_{\mathrm{sim}}(\tilde{\bm{x}}_i, \tilde{\bm{x}}_j, s_{ij})\Big), 
\end{gather}
where $\gamma(t)$ is defined in Equation \ref{eq:gamma_t}. As the training process progresses, the noise rate of the generated similarity labels gradually decreases, thus the weight of $L_{\mathrm{sim}}$ improves gradually.

\subsection{Error Correction}

\noindent We leverage the auxiliary network to alleviate the issue of error accumulation in disambiguation networks. To achieve this, two error correction strategies, i.e., information distillation and confidence refinement, are established, which affect the training of disambiguation network in direct and indirect ways, respectively.

For information distillation, regarding the predicted probability of the auxiliary network as ground-truth distribution, we introduce the following KL divergence-based loss to ensure that the predicted probability of the disambiguation network closely aligns with that of the auxiliary network, 
\begin{gather}
L_{\mathrm{distill}}\left(\bm{x}_i\right) = KL(\operatorname{stop-grad}(\tilde{\bm{p}}_{i}) \lVert \bm{p}_{i}),
\end{gather}
By ignoring the gradients from  $\tilde{\bm{p}}_{i}$ in $L_{\mathrm{distill}}$, we avoid the impact of the disambiguation network's prediction on the auxiliary network.

Additionally, similarly to the confidence calculation in the disambiguation network as described in Equation \ref{eq:cik}, the comprehensive confidence vector $\tilde{\bm{w}}_{i}$ for instance $\tilde{\bm{x}}_i$ is obtained using the prediction results of the auxiliary network. Then, $\tilde{\bm{w}}_{i}$ is utilized to refine the label confidence of the disambiguation network. During the $t$-th epoch, the refined confidence is computed as follows:
\begin{gather}\label{eq:conf_refine}
\hat{\bm{w}}_i(t)=(1-\mu(t)) \bm{w}_i(t)+\mu(t) \tilde{\bm{w}}_i(t),
\end{gather}
where $\mu(t)$ is a non-decreasing function of training epoch $t$. Here, we set $\mu(t)=\min (\rho \times \max (t-t_0, 0), \mu_{\max })$. $t_0$ and $\mu_{\max }$ are hyper-parameters, where $\mu(t)=0$ before  the $t_0$-th training epoch and $0 \leq \mu_{\max }\leq 1$ is the upper bound of $\mu(t)$. The increase speed of $\mu$ depends on $\rho$. During co-training, the original confidence $\bm{w}_i$ in Equation \ref{eq:lrc} is replaced by the refined confidence $\hat{\bm{w}}_i$.

%

 Effective error correction enhances label disambiguation and enables the generation of purer pseudo class labels, further boosting the accuracy of the auxiliary network. The interplay between these two networks forms a virtuous cycle.


\subsection{Training and Inference}

\noindent The overall training loss is as follows:
\begin{gather}
\begin{array}{ll}
L_{\mathrm{total}}\left(\bm{x}_i\right)= L_{\mathrm{disam}}\left(\bm{x}_i\right) + L_{\mathrm{aux}}\left(\tilde{\bm{x}}_i\right) + \gamma(t)L_{\mathrm{distill}}\left(\bm{x}_i\right),
\end{array}
\end{gather}
where the distillation loss based on confidence is also set to the weight $\gamma(t)$.


In the training phase, the disambiguation network is initially warmed up for several epochs to ensure accurate identification of ground-truth labels for certain training instances by the confidence vectors. Subsequently, the pre-trained model parameters are employed to initialize the parameters of the auxiliary network. Moreover, to enhance the efficiency of model training, noisy similarity labels are generated using instances within the same mini batch.

In the inference phase, the predicted probability ensemble of the two learned classifiers naturally enhances performance but incurs additional prediction overhead. Consequently, we opt to utilize only one network for inference. The choice between the disambiguation network and the auxiliary network is insignificant as they ultimately converge to a similar level of performance. In the inference phase, we utilize the disambiguation network for prediction.


\subsection{Complexity Analysis}

\noindent In the training phase, compared to self-training PLL models without co-training, such as DPLL~\cite{12}, AsyCo requires approximately twice the amount of space due to the co-training of two networks. { As for the time complexity, the computational overhead of AsyCo mainly comes from backbone networks, classifiers and loss functions. Due to that backbone networks and classifiers are modular and replaceable in AsyCo, without loss of generality, we denote their computation complexity on an instance as $O(B)$and $O(C)$, respectively. Let the total number of training instances be $N$, in the disambiguation network and auxiliary network, the classification probability of each instance is calculated via the backbone network and classifier once, so the time cost is $O(2NB+2NC)$. The time complexity of RC and CC loss in disambiguation network is $O(3N+3N)$. Since $N^2$ similarity labels are constructed in the auxiliary network, the time complexity of supervised loss is $O(N^2)$. Besides, the time complexity of self-supervised loss in the auxiliary network is $O(3N)$. Totally, the complexity of the AsyCo $O (2NB + 2CN + 6N + N^2 + 3 N) $. Due to that $O (B) \gg O (C)$ and $(B) \gg  O (N) $ in most situations, the final complexity of AsyCo depends on $O(2NB)$, which is almost twice as much as non-co-training models. In practice} , the parallel computing capabilities of GPUs effectively reduce the impact of co-training on training time by leveraging sufficient data parallelism. 

In the inference phase, AsyCo  exhibits comparable time and space complexity to DPLL, which is attributed to the utilization of one single network for prediction. 

\section{Experiments} \label{sec:exp}

\subsection{Experimental Setup}\label{sec:exp-setup}

\subsubsection{Datasets} 

We conduct experiments on {five datasets, which include} three widely used benchmark datasets, including SVHN~\cite{26}, CIFAR-10 and CIFAR-100~\cite{25}, {a text classification dataset CNAE-9 \footnote{ https://archive.ics.uci.edu/dataset/233/cnae+9 }, and a real-world partial label dataset BirdSong, which is for bird song classification task.}   Following~\cite{12}, we construct partially labeled datasets { for the four benchmark datasets}  via two generating processes, i.e., a uniform process and an instance-dependent process. In the uniform generating process, incorrect labels have the same probability $q$ to be a candidate label, where $q$ varies in $\{0.1,0.3,0.5,0.7\}$ on SVHN {, } CIFAR-10, {  and CNAE-9}, and $\{0.01,0.05,0.1,0.2\}$ on CIFAR-100. {  We conduct the instance-dependent candidate generating process on the image datasets.}  We pretrain 18-layer ResNet (ResNet-18) firstly, and the probability of incorrect label $j$ turning into a false positive label is calculated as: 
$$
\frac{g^\prime_j\left(\bm{x}_i\right)}{\max _{k \in Y_i} g^\prime_k\left(\bm{x}_i\right)}, 
$$ where $g_j^\prime(\bm{x}_i)$ is the classification probability into label $j$ calculated by the pretrained ResNet-18 given input $\bm{x}_i$.

\subsubsection{Compared Methods} To evaluate the performance of AsyCo, we choose the following deep PLL methods as competitors: (1) \textbf{CC}~\cite{14}, a classifier-consistent method based on the assumption that candidate label sets are generated uniformly. (2) \textbf{RC}~\cite{14}, a risk-consistent method based on the importance of re-weighting strategy. (3) \textbf{PRODEN}~\cite{15}, a progressive identification method accomplishing classifier learning and label identification simultaneously. (4) \textbf{PiCO}~\cite{11}, a PLL model utilizing contrastive learning module along with a novel class prototype-based label disambiguation algorithm. (5) \textbf{DPLL}~\cite{12}, a model leveraging consistency regularization for deep PLL. (6) \textbf{NCPD}~\cite{21}, a co-training based PLL model employing a progressive disambiguation strategy combined with a network cooperation mechanism for PLL. (7) \textbf{Fully supervised learning}, a model trained with the exact ground-truth labels and cross-entropy loss enhanced by data augmentation.

\subsubsection{Implementation Details} AsyCo is implemented using PyTorch, with an 18-layer ResNet utilized as the backbone network, i.e., serving as the feature encoder { on image datasets. 
Since both CNAE-9 and BirdSong are not very large, we construct linear layers as their feature encoders. For these two non-image datasets, we augment the instances by adding random tokens and Gaussian noise, respectively.} The optimization of the model is carried out using the SGD optimizer, with a momentum value set to 0.9 and a weight decay set to 1e-4. The initial learning rate is set to 0.1, and at the 100th and 150th epochs, the learning rate is divided by 10. The total number of training epochs is set to 200, with warm-up epochs accounting for 50 when $q=0.2$ on CIFAR-100 and 20 in other scenarios. The batch size is set to 64. The value of $\tau$ is searched within the range of $[1,5,10,20,30]$, and ultimately selected as $\tau = 20$. When calculating $\lambda$, the values of $T$ and $\lambda_{max}=1$ are set to 100 and 1, respectively. Furthermore, for the hyper-parameters related to confidence refinement, $\rho$ is set to 0.02, $t_0$ is determined by adding the number of warm-up epochs to 50, and $\mu_{max}$ is set to 0.9. The source code can be found at https://github.com/libeibeics/AsyCo.

For a fair comparison, we employ the same backbone network, learning rate, optimizer, and batch size for all methods, including fully supervised learning. Additionally, for those methods that did not originally employ data augmentation techniques (e.g., RC, CC, and PRODEN), we enhance the models by incorporating the same data augmentation methods used in AsyCo. Particularly,  to ensure PiCO can achieve its optimal performance, we retain data augmentation as described in its original paper. Moreover, we adopt the values of hyper-parameters from their original papers to guarantee that the compared methods are able to achieve their own best performance.  As for training epochs, PiCO undergoes 800 epochs, whereas the remaining models are trained for 200 epochs.  To obtain reliable and robust results, we conduct three repeated experiments with different random seeds and  report  the mean and standard deviation of these results. {All models are trained on GeForce RTX 3090s equipped with 24 GB of memory except for NCPD when setting $q=0.7$ on SVHN. Because there occurs Memory Limitation Error (MLE) in NCPD on the SVHN data set when q=0.7, we run the experiment on V100 with 32 GB of memory to get the results.} 

\subsection{Performance Comparison}

\begin{table*}[t]
\footnotesize
\caption{Accuracy (mean$\pm$std) comparison with uniform partial labels on different ambiguity levels on image datasets. (The best performance in each column is highlighted in bold. The second best performance in each column is underlined. The improvement is calculated based on the best competitor.)}
\label{tab:performance-comparison}
\begin{center}
\begin{tabular}{c|ccccc}
\toprule
Datasets & Models    & $q$ = 0.1   & $q$ = 0.3 & $q$ = 0.5 & $q$ = 0.7 \\ 
\midrule
\multirow{9}{*}{SVHN} 
    & \cellcolor{mygray} Fully Supervised & \multicolumn{4}{c}{\cellcolor{mygray} 97.435 $\pm$ 0.016\%}\\
    & CC & 97.348 $\pm$ 0.100\% & 97.139 $\pm$ 0.048\% & 96.978 $\pm$ 0.020\% & \underline{96.377 $\pm$ 0.020\%}\\
    & RC & 97.292 $\pm$ 0.085\% & 97.243 $\pm$ 0.128\% & 97.050 $\pm$ 0.049\% & 95.898 $\pm$ 0.108\%\\
    & PRODEN & 97.081 $\pm$ 0.077\% & 96.445 $\pm$ 0.290\% & 96.183 $\pm$ 0.325\% & 94.573 $\pm$ 0.492\% \\
    & PiCO & 95.680 $\pm$ 0.080\% & 95.585 $\pm$ 0.015\% & 95.630 $\pm$ 0.020\% & 95.150 $\pm$ 0.024\% \\
    & DPLL & 97.261 $\pm$ 0.029\% & 97.062 $\pm$  0.013\% & 96.797 $\pm$ 0.033\% & 94.972 $\pm$ 0.106\% \\
    & NCPD & \textbf{97.469 $\pm$ 0.011\%} & \underline{97.431 $\pm$ 0.045\%} & \underline{97.325 $\pm$ 0.041\%} &  18.865 $\pm$ 2.157\% \\
    & \cellcolor{mygray}  AsyCo & \cellcolor{mygray} \underline{97.374 $\pm$ 0.015\%} & \cellcolor{mygray} \textbf{97.471 $\pm$ 0.086\%} & \cellcolor{mygray} \textbf{97.553 $\pm$ 0.023\%} & \cellcolor{mygray} \textbf{97.539 $\pm$ 0.013\%} \\
    & \cellcolor{mygray} Improv. &  \cellcolor{mygray}\textbf{-} & \cellcolor{mygray}$\uparrow$ \textbf{0.040\%} & \cellcolor{mygray}$\uparrow$ \textbf{0.228\%} & \cellcolor{mygray}$\uparrow$ \textbf{1.162\%} \\
\midrule
Datasets & Models    & $q$ = 0.1   & $q$ = 0.3 & $q$ = 0.5 & $q$ = 0.7 \\ 
\midrule
\multirow{9}{*}{CIFAR-10} 
    & \cellcolor{mygray}  Fully Supervised & \multicolumn{4}{c}{\cellcolor{mygray} 96.458 $\pm$ 0.062\%}\\
    & CC & 94.129 $\pm$ 0.181\%  & 93.226 $\pm$ 0.261\% & 92.102 $\pm$ 0.155\%  & 88.846 $\pm$ 0.031\% \\
    & RC & 94.950 $\pm$ 0.100\% & 94.610 $\pm$ 0.054\% & 94.139 $\pm$ 0.059\% & 92.423 $\pm$ 0.051\% \\
    & PRODEN & 94.443 $\pm$ 0.213\% & 93.845 $\pm$ 0.326\% & 93.466 $\pm$ 0.243\% & 91.259 $\pm$ 0.780\% \\
    & PiCO & 94.357 $\pm$ 0.109\% & 94.183 $\pm$ 0.179\% & 93.697 $\pm$ 0.238\% & 92.157 $\pm$ 0.209\% \\
    & DPLL & 95.905 $\pm$ 0.052\% & \underline{95.654 $\pm$ 0.208\%} & \underline{95.365 $\pm$ 0.140\%} & \underline{93.856 $\pm$ 0.366\%} \\
    & NCPD & \underline{96.284 $\pm$ 0.050\%} & 95.280 $\pm$ 0.110\% & 95.280 $\pm$ 0.110\% & 76.583 $\pm$ 0.522\%\\
    & \cellcolor{mygray}  AsyCo & \cellcolor{mygray} \textbf{96.645 $\pm$ 0.004\%} & \cellcolor{mygray} \textbf{96.279 $\pm$ 0.030\%} & \cellcolor{mygray} \textbf{96.003 $\pm$ 0.013\%} & \cellcolor{mygray} \textbf{95.550 $\pm$ 0.007\%}\\
    & \cellcolor{mygray}Improv. & \cellcolor{mygray}$\uparrow$ \textbf{ 0.361\%} & \cellcolor{mygray}$\uparrow$ \textbf{0.625\%} & \cellcolor{mygray}$\uparrow$ \textbf{0.638\%} & \cellcolor{mygray}$\uparrow$ \cellcolor{mygray}\textbf{1.694\%} \\
\midrule
Datasets & Models    & $q$ = 0.01   & $q$ = 0.05 & $q$ = 0.1 & $q$ = 0.2 \\ 
\midrule
\multirow{9}{*}{CIFAR-100} 
    & \cellcolor{mygray} Fully Supervised & \multicolumn{4}{c}{\cellcolor{mygray} 80.385 $\pm$ 0.013\%}\\
    & CC & 75.560 $\pm$ 0.537\% & 75.138 $\pm$ 0.154\% & 73.224 $\pm$ 1.017\% & 69.035 $\pm$ 0.339\%\\
    & RC & 76.252 $\pm$ 0.168\% & 75.689 $\pm$ 0.129\% & 74.737 $\pm$ 0.282\% & 72.708 $\pm$ 0.358\%\\
    & PRODEN & 76.147 $\pm$ 0.291\% & 75.682 $\pm$ 0.097\% & 74.604 $\pm$ 0.285\% & 72.512 $\pm$ 0.212\%\\
    & PiCO & 73.145 $\pm$ 0.035\% & 72.585 $\pm$ 0.145\% & 59.365 $\pm$ 0.445\% & 25.545 $\pm$ 0.715\%\\
    & DPLL & \underline{79.300 $\pm$ 0.262\%} & \underline{78.855 $\pm$ 0.165\%} & \underline{78.064 $\pm$ 0.050\%} & \underline{76.316 $\pm$ 0.232\%}\\
    & NCPD & 78.190 $\pm$ 0.080\% & 76.990 $\pm$ 0.041\% & 71.923 $\pm$ 0.042\% & 42.701 $\pm$ 0.832\%  \\
    & \cellcolor{mygray}  AsyCo & \cellcolor{mygray} \textbf{80.775 $\pm$ 0.010\%} & \cellcolor{mygray} \textbf{80.433 $\pm$ 0.087\%} & \cellcolor{mygray} \textbf{79.668 $\pm$ 0.058\%} & \cellcolor{mygray} \textbf{78.061 $\pm$ 0.001\%} \\
    & \cellcolor{mygray}Improv. &  \cellcolor{mygray}$\uparrow$ \textbf{1.475\%} & \cellcolor{mygray}$\uparrow$ \textbf{1.578\%} & \cellcolor{mygray}$\uparrow$ \textbf{1.604\%} & \cellcolor{mygray}$\uparrow$ \textbf{1.745\%} \\    
\bottomrule
\end{tabular}
\end{center}
\end{table*}

\begin{table}[t]
\footnotesize
\caption{ Accuracy comparison to supervised learning without data augmentation.}
\label{tab:cmp-sup-withoutda}
\begin{center}
\begin{tabular}{c|ccc}
\toprule
Datasets & SVHN & CIFAR-10 & CIFAR-100 \\ 
\midrule
Fully Supervised w/o D.A. & 95.647 $\pm$ 0.147\% &  \textbf{95.210 $\pm$ 0.245\%} & 75.621 $\pm$ 0.478\% \\
AsyCo w/o  D.A. & \textbf{96.338 $\pm$ 0.030\%}($q$ = 0.1)& 94.981 $\pm$ 0.050\% ($q$ = 0.1) &  \textbf{76.415 $\pm$ 0.175\%} ($q$ = 0.01)\\
\bottomrule
\end{tabular}
\end{center}
\end{table}

\begin{table*}[t]
\footnotesize
\caption{Accuracy (mean$\pm$std) comparison with uniform partial labels on different ambiguity levels on CNAE-9. }
\label{tab:performance-cmp-cnae}
\begin{center}
\begin{tabular}{c|ccccc}
\toprule
Datasets & Models    & $q$ = 0.1   & $q$ = 0.3 & $q$ = 0.5 & $q$ = 0.7 \\ 
\midrule
\multirow{9}{*}{ CNAE-9} 
    & \cellcolor{mygray}  Fully Supervised & \multicolumn{4}{c}{\cellcolor{mygray} 95.095 $\pm$ 1.044\%}\\
    & CC & 93.673 $\pm$ 0.218\% & 92.438 $\pm$ 0.437\% &  91.204 $\pm$ 0.378\% &  82.870 $\pm$ 0.655\%\\
    & RC &  92.901 $\pm$ 0.218\% &  91.870 $\pm$ 0.838\% &  89.043 $\pm$ 0.787\% &  82.870 $\pm$ 0.655\%\\
    & PRODEN &  93.673 $\pm$ 0.873\% &  \underline{93.827 $\pm$ 0.436}\% &  90.278 $\pm$ 0.378\% &  79.907 $\pm$ 0.786\%\\
    & PiCO &  91.665 $\pm$ 0.465\%  &  86.340 $\pm$ 0.230\% &  57.713 $\pm$ 2.838\% &  44.940 $\pm$ 2.008\%\\
    & DPLL &  \textbf{95.296	$\pm$  0.074\%} &  93.210 $\pm$ 0.436\% &  \underline{92.901 $\pm$ 1.431\%} &  82.716 $\pm$ 1.091\%\\
    & NCPD &  \underline{95.139 $\pm$ 0.694\%} &  93.288 $\pm$ 0.232\% &  92.439 $\pm$ 0.952\% &  \underline{84.491 $\pm$ 0.231\%}  \\
    & \cellcolor{mygray}  AsyCo & \cellcolor{mygray}  95.062 $\pm$ 0.437\% & \cellcolor{mygray}  \textbf{93.982 $\pm$ 0.756\%} & \cellcolor{mygray}  \textbf{93.235 $\pm$ 0.948\%} & \cellcolor{mygray}  \textbf{86.728 $\pm$ 0.787\%} \\
    & \cellcolor{mygray}  Improv. &  \cellcolor{mygray} \textbf{-} & \cellcolor{mygray}$\uparrow$  \textbf{0.155\%} & \cellcolor{mygray}$\uparrow$ \textbf{0.334\%} & \cellcolor{mygray}$\uparrow$ \textbf{2.237\%} \\
    
\bottomrule
\end{tabular}
\end{center}
\end{table*}

\begin{table}[t]
    \footnotesize
    \centering
    \begin{minipage}[t]{0.4\textwidth}
    \caption{ Accuracy (mean$\pm$std) comparison on the real-world dataset BirdSong. }
    \label{tab:performance-cmp-birdsong}
    \tabcolsep 10pt
    \begin{center}
    \begin{tabular}{c|c}
    \toprule
     Models    &  BirdSong\\ 
    \midrule
     CC &  71.420 $\pm$ 0.900\%  \\
     RC &  70.263 $\pm$ 0.274\%  \\
     PRODEN &  70.623 $\pm$ 0.845\% \\
     PiCO &  71.700 $\pm$ 0.800\%  \\
     DPLL  &  \underline{72.093 $\pm$ 0.285\%}  \\
     NCPD  &    66.960 $\pm$ 1.100\%  \\
    \cellcolor{mygray} AsyCo &  \cellcolor{mygray}  \textbf{72.770 $\pm$ 0.070\%}\\
    \midrule
    \cellcolor{mygray} Improv. & \cellcolor{mygray}  $\uparrow$ \textbf{0.677\%} \\
    \bottomrule
    \end{tabular}
    \end{center}
\end{minipage}
\hspace{0.08\textwidth} 
\begin{minipage}[t]{0.45\textwidth}
    \caption{Accuracy (mean$\pm$std) comparison on CIFAR-10 and SVHN with instance-dependent partial labels. }
    \label{tab:performance-instance}
    \begin{center}
    \begin{tabular}{c|ccccc}
    \toprule
    Models    & SVHN    & CIFAR-10\\ 
    \midrule
    CC & 96.072  $\pm$ 0.041\% & 93.701 $\pm$ 0.006\%  \\
    RC & \underline{96.899  $\pm$ 0.087\%} & 93.270  $\pm$ 0.013\%  \\
    PRODEN & 95.626  $\pm$ 0.084\% & 92.409  $\pm$ 0.041\% \\
    PiCO & 95.615  $\pm$ 0.045\% & 92.715  $\pm$ 0.055\%  \\
    DPLL  & 95.796  $\pm$ 0.015\%  & 93.657  $\pm$ 0.104\%  \\
    NCPD  &   96.633 $\pm$ 0.056\%  &  \underline{94.011 $\pm$ 0.011\%}  \\
    \cellcolor{mygray}AsyCo &  \cellcolor{mygray} \textbf{97.528 $\pm$ 0.008\%} & \cellcolor{mygray}\textbf{95.301 $\pm$ 0.046\%}   \\
    \midrule
    \cellcolor{mygray}Improv. & \cellcolor{mygray} $\uparrow$ \textbf{0.895\%} & \cellcolor{mygray} $\uparrow$ \textbf{1.290\%}     \\
    \bottomrule
    \end{tabular}
    \end{center}
    \end{minipage}
\end{table}

\noindent The performance comparison results are shown in Table \ref{tab:performance-comparison}. From the results, we have following observation and analysis.

Firstly, DPLL, which is the state-of-the-art deep PLL model, performs differently across datasets. Specifically, in the experiments conducted on CIFAR-10 and CIFAR-100, it is observed that the DPLL model significantly outperforms other compared models, which highlights the effectiveness of manifold consistency regularization in enhancing classification accuracy. But on SVHN, DPLL is slightly inferior to other comparison methods such as PRODEN, RC, CC. The discrepancy in performance on different datasets of DPLL may result from the nature of the task at hand. SVHN involves a relatively simpler classification task of recognizing 0-9 digits, whereas the image classification task on CIFAR datasets is more complex. Due to the inherent simplicity of the classification task on SVHN, the task can be well-solved by simple models, e.g, RC and CC.

Secondly, the NCPD model, which is based on symmetric co-training, exhibits strong performance when $q$ is small while significantly deteriorates as $q$ increases. For instance, it outperforms all other methods when $q=0.1$ on CIFAR-10 and even slightly surpasses AsyCo at the same $q$ value on SVHN. However, when $q=0.7$ on both CIFAR-10 and SVHN, NCPD performs the worst compared to its competitors. This decline in performance can be attributed to the fact that NCPD converts the partially labeled dataset into a noisy dataset through multi-birth duplication. Consequently, the noise rate of the generated dataset is dependent on the average number of candidate labels, meaning that a higher $q$ leads to a greater noise rate. As a result, larger values of $q$ result in more severe degradation of performance. Additionally, NCPD suffers from another limitation, that is, larger values of $q$ lead to an increased number of instances in the generated noisy dataset, resulting in higher time and space overhead for model training.

Thirdly, in general, the proposed AsyCo demonstrates significant superiority over all its competitors on three datasets with various $q$ values. For instance, on CIFAR-10, when $p$ takes on the values of 0.1, 0.3, 0.5, and 0.7, the performance of AsyCo surpasses that of the best competitor by 0.361\%, 0.625\%, 0.638\%, and 1.694\%, respectively. This clearly indicates the remarkable performance of AsyCo. Moreover, the effectiveness of asymmetric dual-task co-training is convincingly demonstrated by the superiority of AsyCo over NCPD. Additionally, it is noteworthy that the accuracy of AsyCo remains highly competitive compared to supervised learning.
{  When $q$ is small, for example, on CIFAR-10 with $q=0.1$, as well as on CIFAR-100 with $q=0.01$ and $q=0.05$. AsyCo achieves even better performance than supervised learning in certain scenarios.} { We futher remove data augmentation from both AsyCo and supervised learning and compare their performance. As shown in Table \ref{tab:cmp-sup-withoutda},   without data augmentation, AsyCo still outperforms fully supervised learningsignificantly, which suggests that AsyCo effectively incorporates different capabilities of two networks through co-training, which can mine high-quality supervised signals from partially labeled data, for example, the relationship between instance pairs.}

Furthermore, AsyCo exhibits strong robustness with respect to data quality degradation. Specifically, as the value of $q$ increases, the performance of most compared methods noticeably deteriorates. For instance, on CIFAR-10, the accuracy of the competitors decreases by a range of 2\% to 20\% as $q$ increases from 0.1 to 0.7. In contrast, the accuracy of AsyCo only fluctuates within the narrower range of 96.645\% to 95.550\%. Consequently, the disparity in accuracy between the top-performing competitor and AsyCo becomes more pronounced with increasing values of $q$. Specifically, on SVHN, the performance improvements achieved by AsyCo are 0.228\% and 1.162\% for $q=0.5$ and $q=0.7$, respectively. Similarly, on CIFAR-100, the accuracy improvements for AsyCo at $q$ values of 0.01 and 0.2 are 1.475\% and 1.745\%, respectively.

{ As shown in Table \ref{tab:performance-cmp-cnae} and Table \ref{tab:performance-cmp-birdsong}, on the two non-image datasets, i.e., CNVAE-9 and BirdSong, AsyCo still outperforms all the competitors in most situations. In addition, on CNVAE-9, AsyCo also shows good robustness as the $q$ increases, and the accuracy advantage over the competitors becomes more and more obvious, which is the same as that on the image dataset. This shows that AsyCo can be well generalized to datasets in different domains.}

Last but not least, in addition to the outstanding performance on the uniformly generated partially labeled datasets, as shown in Table \ref{tab:performance-instance}, AsyCo outperforms other models when applied to instance-dependent partially labeled datasets. It exhibits a notable performance increase of 1.290\% on CIFAR-10 and achieves an accuracy improvement of 0.895\% on SVHN, validating the capabilities and generalizability of AsyCo.

\subsection{Ablation Study}
\subsubsection{Impact of the Auxiliary Network}\label{sec:impact-auxnet}
\begin{table}[t]
\footnotesize
\caption{Ablation study of co-training, where the disambiguation network is trained individually.}
\label{tab:performance-impact-auxnet}
\tabcolsep 12pt
\begin{center}
\begin{tabular}{c|ccc}
\toprule
Datasets & $q$ = 0.3 & $q$ = 0.5 & $q$ = 0.7 \\ 
\midrule
SVHN &	97.323 $\pm$ 0.021\% ($\downarrow$ \textbf{0.148\%}) &	97.154 $\pm$ 0.090\% ($\downarrow$ \textbf{0.399\%}) &	96.238 $\pm$ 0.069\% ($\downarrow$ \textbf{1.301\%})  \\
\midrule
Datasets & $q$ = 0.3 & $q$ = 0.5 & $q$ = 0.7 \\ 
\midrule
CIFAR-10 &  96.191 $\pm$ 0.077\% ($\downarrow$ \textbf{0.088\%}) & 95.746 $\pm$ 0.066\% ($\downarrow$ \textbf{0.257\%}) & 94.455 $\pm$ 0.102\% ($\downarrow$ \textbf{1.095\%})\\
\midrule
Datasets & $q$ = 0.05 & $q$ = 0.1 & $q$ = 0.2 \\ 
\midrule
CIFAR-100 &79.969 $\pm$  0.136\% ($\downarrow$ \textbf{0.464\%}) &	79.263 $\pm$  0.056\% ($\downarrow$ \textbf{0.405\%}) &	77.921 $\pm$  0.020\% ($\downarrow$ \textbf{0.140\%}) \\
\bottomrule
\end{tabular}
\end{center}
\end{table}

\noindent In order to examine the impact of the auxiliary network, we conduct an experiment by excluding the auxiliary network from AsyCo and training the disambiguation network separately. This setting is equivalent to directly removing the error correction module from AsyCo, thereby eliminating the impact of the auxiliary network on the disambiguation network. The experimental results, shown in Table \ref{tab:performance-impact-auxnet}, demonstrate a decrease in performance across all datasets and $p$ values, indicating that co-training can enhance the prediction accuracy of PLL. Particularly, AsyCo exhibits a more pronounced advantage as the value of $q$ increases on CIFAR-10 and SVHN, providing further evidence that co-training enhances the robustness of the model.

\subsubsection{Impact of Asymmetric Co-training Architecture}


\begin{table}[t]
    \scriptsize
    \caption{Impact of asymmetric co-training, where model variant SyCo is trained based on symmetric co-training strategy, that is, the two networks are both trained by PLL tasks.}
    \label{tab:performance-pllpll}
    \tabcolsep 12pt
    \begin{center}
    \begin{tabular}{c|ccc}
    \toprule
    Settings & SVHN, $q=0.7$ & CIFAR-10, $q=0.7$ & CIFAR-100, $q=0.2$ \\
    \midrule
    \cellcolor{mygray}AsyCo & \cellcolor{mygray}\textbf{97.539 $\pm$ 0.013\%}  & \cellcolor{mygray}\textbf{95.550 $\pm$ 0.007\%} & \cellcolor{mygray}\textbf{78.061 $\pm$ 0.001\%}\\
    \multirow{2}{*}{SyCo (Symmetric co-training model variant)} & 96.584 $\pm$ 0.066\%  & 94.733 $\pm$ 0.210\%  & 78.028 $\pm$ 0.004\%  \\
    & \textbf{$\downarrow$ 0.955\%} & \textbf{$\downarrow$ 0.607\%}& \textbf{$\downarrow$ 0.033\%}\\
    \bottomrule
    \end{tabular}
    \end{center}
\end{table}

\noindent Different from existing co-training PLL models, AsyCo is built upon asymmetric co-training architecture. In this section, we construct a symmetric co-training model variant called SyCo to explore the impact of the asymmetric co-training architecture. In SyCo, the two networks are initialized differently and both trained with the same PLL loss as Equation \ref{eq:ldisam}. Additionally, SyCo employs a symmetric KL divergence-based distillation loss to enable the two networks to interact with each other for error correction. The performance comparison between SyCo and AsyCo is presented in Table \ref{tab:performance-pllpll}, which reveals a significant decrease in accuracy, particularly on CIFAR-10 and SVHN. This finding demonstrates that the asymmetric dual-task co-training employed in AsyCo is more effective compared to traditional symmetric co-training techniques. The underlying reason can be the fact that training the two networks with distinct tasks compels them to explicitly learn from different perspectives. Therefore, the two networks have a higher likelihood of acquiring complementary information, thereby avoid the error accumulation via communicating with each other.

\subsubsection{Impact of Error Correction Strategy}


    \begin{table}[t]
        \footnotesize
        \caption{Ablation study of error correction strategies.}
        \label{tab:ablation-errorcorrection}
        \tabcolsep 10pt
        \begin{center}
        \begin{tabular}{c|cc}
        \toprule
        Settings & SVHN, $q=0.7$ & CIFAR-10, $q=0.7$ \\
        \midrule
        \cellcolor{mygray}AsyCo & \cellcolor{mygray}\textbf{97.539 $\pm$ 0.013\%} & \cellcolor{mygray}\textbf{95.550 $\pm$ 0.007\%}  \\
        w/o distillation & 97.350 $\pm$ 0.053\% (\textbf{$\downarrow$ 0.189\%}) & 95.400 $\pm$ 0.030\% (\textbf{$\downarrow$ 0.150\%}) \\
        w/o confidence refinement & 97.481 $\pm$ 0.057\% (\textbf{$\downarrow$ 0.058\%}) & 95.456 $\pm$ 0.062\% (\textbf{$\downarrow$ 0.094\%}) \\
        \bottomrule
        \end{tabular}
        \end{center}
    \end{table}

We perform an ablation study of the two error correction strategies, namely distillation and confidence refinement. The experimental results are presented in Table \ref{tab:ablation-errorcorrection}. It is evident that removing either strategy leads to a slight decrease in accuracy, which indicates that the combination of them plays a more substantial role. Refer to that the significant performance decrease caused by removing the auxiliary network, i.e., removing the whole error correction module shown in Table \ref{tab:performance-impact-auxnet}, it can be inferred that either of the proposed error correction strategies contributes to the final classification accuracy. Moreover, removing distillation results in a larger decline in performance compared to removing confidence refinement. This may result from the fact that distillation facilitates a more direct and timely influence of disambiguation models on auxiliary models.

\subsubsection{Impact of Label Transformation}



\begin{table}[t]
    \footnotesize

    \caption{Prediction accuracy when the auxiliary network is trained based on the pseudo class labels.}
    \label{tab:pairwise-ce}
    \tabcolsep 6pt
    \begin{center}
    \begin{tabular}{c|ccc}
    \toprule
    Settings & SVHN, $q=0.7$ & CIFAR-10, $q=0.7$ & CIFAR-100, $q=0.2$ \\
    \midrule
    \cellcolor{mygray}AsyCo & \cellcolor{mygray}\textbf{97.539 $\pm$ 0.013\%}  & \cellcolor{mygray}\textbf{95.550 $\pm$ 0.007\%} & \cellcolor{mygray}\textbf{78.061 $\pm$ 0.001\%}\\
    \multirow{2}{*}{AsyCo trained with the pseudo class labels} & 97.380 $\pm$ 0.010\%  & 95.164 $\pm$ 0.015\%  & 77.489 $\pm$ 0.053\%  \\
    & \textbf{$\downarrow$ 0.159 \%} & \textbf{$\downarrow$ 0.386 \%} & \textbf{$\downarrow$ 0.572 \%}\\
    \bottomrule
    \end{tabular}
    \end{center}
    \end{table}

\begin{figure}[!t]
    \begin{minipage}[c]{0.65\textwidth}
    \centering
    \begin{minipage}[c]{0.46\textwidth}
    \centering
    \includegraphics[width=\textwidth]{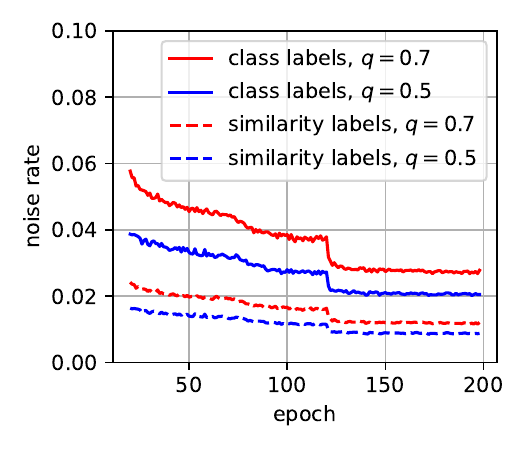}
    \end{minipage}
    \hspace{0.02\textwidth}
    \begin{minipage}[c]{0.46\textwidth}
    \centering
    \includegraphics[width=\textwidth]{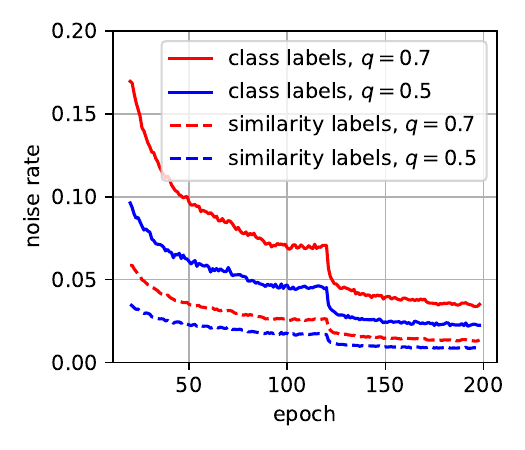}
    \end{minipage}
    \caption{Comparison in terms of noise rate between pairwise similarity labels and pseudo class labels during  training AsyCo. (a) Noise rates on SVHN; (b) Noise rates on CIFAR-10.}
    \label{fig:noise-rate-comparison}
\end{minipage}
\hspace{0.02\textwidth}
\begin{minipage}[c]{0.3\textwidth}
    \centering
    \includegraphics[width=\textwidth]{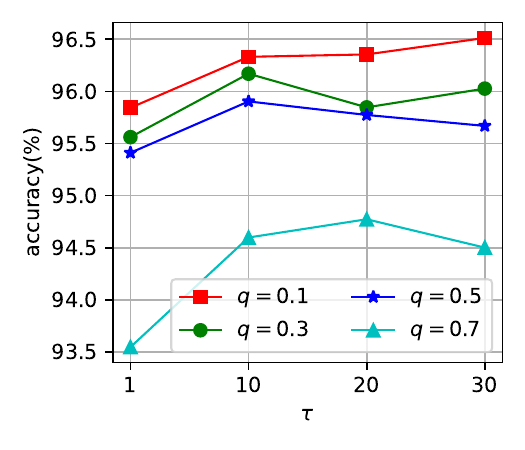}
    \caption{Impact of the temperature parameter $\tau$ on prediction accuracy on CIFAR-10.}
    \label{fig:hyper:tau}
    \end{minipage}

    \end{figure}

The impact of converting pseudo class labels into pairwise similarity labels is analyzed from two aspects. Firstly, a comparison is conducted between the noise rates of pairwise similarity labels and pseudo class labels while optimizing AsyCo, as depicted in Figure \ref{fig:noise-rate-comparison}. It is obvious that the noise rate of similarity labels is significantly lower than that of the noisy class labels. Secondly, a model variant of AsyCo is constructed that disregards label transformation and instead trains the auxiliary network directly using pseudo class labels. The degradation in performance, as demonstrated in Table \ref{tab:pairwise-ce}, clearly illustrates that transforming noisy class labels into noisy pairwise similarity labels  reduces the influence of noise labels on prediction accuracy.

\subsubsection{Ablation Study of Data Augmentation}

    \begin{table}[t]
        \footnotesize
        \caption{Ablation study of data augmentation.}
        \label{tab:abl-da}
        \tabcolsep 12pt
        \begin{center}
        \begin{tabular}{l|ccc}
        \toprule
        Settings & SVHN, $q$ = 0.7 & CIFAR-10, $q$ = 0.7 & CIFAR-100, $q$ = 0.2 \\ 
        \midrule
        w/o D.A. & 96.290 $\pm$ 0.019\%  & 93.093 $\pm$ 0.001\% & 74.627 $\pm$ 0.195\% \\
        w/  1 D.A. &  \textbf{97.705 $\pm$ 0.001\%}  & 95.475 $\pm$ 0.033\%  & 77.844 $\pm$ 0.350\% \\
        w/  2 D.A. (the original AsyCo)	 &  97.539 $\pm$ 0.013\%  &  \textbf{95.550 $\pm$ 0.007\%}  &  \textbf{78.061 $\pm$ 0.001\%} \\
        w/  3 D.A. & 97.166 $\pm$ 0.262\%  & 95.415 $\pm$ 0.017\%  & 77.985 $\pm$ 0.065\% \\
        \bottomrule
        \end{tabular}
        \end{center}
        \end{table}

{ In order to investigate the impact of data augmentation, we construct model variants based on AsyCo by removing data augmentation and varying the number of data augmentations per instance, as shown in Table \ref{tab:abl-da}. The experimental results show that removing data augmentation or over-increasing the number of times each sample is augmented, e.g., 3 augmentations per sample, brings about a decrease in classification accuracy. Generally, the model performs best on the CIFAR datasets with two data augmentations for each instance. This experiment illustrates the contribution of data augmentation to classification accuracy and the importance of choosing a proper number of data augmentations.}

\subsection{Further Analysis}

\subsubsection{Impact of Backbone Network}

\begin{table}[t]
\footnotesize
\tabcolsep 20pt
\caption{Impact of backbone networks, where ResNet-18 is replaced with Wide-ResNet-34-10.}
\label{tab:impact-backbone}
\begin{center}
\begin{tabular}{l|ccc}
\toprule
%


Settings & SVHN, $q$ = 0.7 & CIFAR-10, $q$ = 0.7 \\ 
\midrule
DPLL & 96.627 $\pm$ 0.115 ($\uparrow$ \textbf{1.655\%}) & 94.910 $\pm$ 0.060 ($\uparrow$ \textbf{2.753\%}) \\
\cellcolor{mygray}AsyCo &  \cellcolor{mygray}\textbf{97.941 $\pm$ 0.104 ($\uparrow$ \textbf{0.402 \%})} &  \cellcolor{mygray}\textbf{96.643 $\pm$ 0.023 ($\uparrow$\textbf{1.093 \%})} \\
\bottomrule
\end{tabular}
\end{center}
\end{table}

We replace the backbone network ResNet-18 with Wide-ResNet-34-10 in order to analyze the impact of the backbone network on performance. The results presented in Table \ref{tab:impact-backbone} demonstrate that a stronger backbone network can lead to additional performance improvement in AsyCo. Specifically, when ResNet-18 is replaced with Wide-ResNet, we observe accuracy improvements of approximately 0.402 \% and 1.093\% on SVHN and CIFAR-10, respectively. These findings indicate the substantial performance potentials of AsyCo. { As a comparison, we also investigate the performance of replacing the backbone network for the competitors. And due to space limitations, we list the accuracy of the strongest competitor, i.e., DPLL, on the CIFAR dataset after adopting wide-resnet as its backbone network. It can be seen that  a stronger backbone network leads DPLL to a significant performance improvement, though the performance of DPLL is still significantly behind on the SVHN and CIFAR-10 datasets compared to AsyCo.}


\subsubsection{Impact of Temperature Parameter $\tau$}


\noindent The temperature coefficient $\tau$ in Equation \ref{eq:prob_ik} determines the smoothness of the classification probability. Here, we study the effect of $\tau$ on the disambiguation network. As shown in Figure \ref{fig:hyper:tau}, the accuracy of the disambiguation network is visualized when different values of $\tau$ (1, 10, 20, and 30) are applied, where the disambiguation network is trained individually. It is observed that an improvement in performance is evident when $\tau>1$, regardless of the $q$ values. The reason is that properly smoothed confidence avoids aggressive optimization and reflects detailed information of instances, e.g, the correlation between the instance and false positive labels. However, excessively smooth confidence resulting from a very large value of $\tau$ leads to a degradation in performance. This is due to the inability to clearly differentiate the weights of each candidate label.


\section{Related Work}

\subsection{Traditional Partial-label Learning}

\noindent Traditional PLL methods can be divided into two categories, i.e., average-based methods and identification-based methods. The average-based methods treat each label in a candidate label set equally~\cite{27,2,28,disambiguation_free}.  However, the ground truth label of each instance is easily overwhelmed, especially when the number of candidate labels is large. To alleviate the problem, identification-based methods try to disambiguate ground-truth label from candidate label sets. Some of them utilize a two-phase strategy~\cite{28}, i.e., first refining label confidence, then learning the classifier, while others progressively refine confidence during learning the classifier~\cite{5}. Besides, manifold consistency regularization, which assumes that similar instances are supposed to have similar label distributions, has been widely employed in PLL to estimate the label confidence and learn the classifier simultaneously~\cite{28,29,5,6}. Recently, an algorithm towards multi-instance PLL problem has been explored~\cite{multi_instance_pll}, which assumes each training sample is associated with not only multiple instances but also a candidate label set that contains one ground-truth label and some false positive labels. However, these traditional methods are usually linear or kernel-based models, which are hard to deal with large-scale datasets.

\subsection{Deep Partial-label Learning}

\noindent  With the powerful modeling capability of deep learning, deep PLL methods can handle high-dimensional features and outperform traditional methods.  Assuming a uniform set-level partial label generation process, Feng et al.~\cite{14} propose a classifier-consistent loss, which is model-agnostic and can be directly combined with deep classifiers and variant optimizers. However, it treats each candidate label equally. RC~\cite{14}, PRODEN~\cite{15} and LWS ~\cite{13} leverage  self-training and  estimate label confidence and train the model with it iteratively. PiCO~\cite{11} and DPLL~\cite{12} further explore contrastive learning and manifold consistency in self-training PLL models, respectively. Nevertheless, self-training PLL models suffer from error accumulation problem resulted from mistakenly disambiguated instances. To address this issue, NCPD~\cite{21} converts the patial labels  into noisy labels via multi-birth duplication and adopts a typical co-training NLL method called co-teaching~\cite{20}. Unfortunately, the label transformation in NCPD results in a high noise rate, limiting classification accuracy and resulting high time and space complexity. Besides, the two networks in NCPD are trained with the same input data and loss functions and easily reach a consensus, thereby cannot correct errors for each other effectively, which naturally motivates us to improve them in our research. 

\vspace{-0.3cm}
\section{Conclusion}
\vspace{-0.2cm}

\noindent In this paper,  we propose an asymmetric dual-task co-training PLL model AsyCo, which  forces them learn from different views explicitly by training  a  disambiguation network and an auxiliary network via optimizing different tasks.  To alleviate the error accumulation problem of self-training PLL models, we establish an information flow loop between the two networks in AsyCo as their collaboration mechanism, i.e., the disambiguation network provides the auxiliary network with the identified pseudo class labels, while the auxiliary network conducts error correction for the disambiguation network through distillation and confidence refinement. Results of experiments on benchmark datasets fully demonstrate the superior performance of AsyCo compared to existing PLL models and the effectiveness of asymmetric co-training in error elimination. { Though AsyCo achieves excellent performance, it also has limitations. Like other co-training-based models, it requires almost twice the computational space to complete the training, which brings higher training overhead. In the future, }we will further conduct research on different co-training architectures and network cooperation mechanisms to tap the potential of dual-task co-training models for PLL. 






\Acknowledgements{This work is supported by the National Natural Science Foundation of China (Grant No. 62106028) and Chongqing Overseas Chinese Entrepreneurship and Innovation Support Program. Besides, this work is also supported by the National Natural Science Foundation of China (Grant No. 62072450) and Chongqing Science and Technology Bureau (CSTB2022TIAD-KPX0180). The authors wish to thank the associate editor and anonymous reviewers for their helpful comments and suggestions.}






\end{document}